\documentclass[runningheads]{llncs}
\usepackage{ragged2e}
\usepackage{url}
\usepackage{booktabs, makecell, tabularx}

\setcellgapes{3pt}
\newcolumntype{L}{>{\RaggedRight}X}
\newcolumntype{P}[1]{>{\raggedright\arraybackslash}p{#1}}
\usepackage{caption}
\usepackage[dvipsnames]{xcolor}
\usepackage[toc,page]{appendix} 
\usepackage{graphicx,booktabs}
\usepackage[figuresright]{rotating}

\usepackage[verbose]{placeins}
\usepackage{blindtext}
\usepackage[T1]{fontenc}
\usepackage{enumitem}
\usepackage{float}
\usepackage{soul}
\renewcommand\hl[1]{#1}
\begin{document}
\title{Large Models in Dialogue for Active Perception and Anomaly Detection 
}
%
%
\author{Authors' Response Letter}
\title{Large Models in Dialogue for Active Perception and Anomaly Detection 
}
%
%
\author{Tzoulio Chamiti\inst{1}
\and Nikolaos Passalis\inst{2}
\and Anastasios Tefas\inst{1}} 
\authorrunning{T. Chamiti et al.}
%
\institute{Computational Intelligence and Deep Learning Group, AIIA Lab.,
\\Dept. of Informatics \\
\and
Dept. of Chemical Engineering\\
Aristotle University of Thessaloniki, Thessaloniki 541 24, Greece
\email{t.chamiti@csd.auth.gr, passalis@auth.gr, tefas@csd.auth.gr}\\
}
\maketitle              
\begin{abstract}
Autonomous aerial monitoring is an important task aimed at gathering information from areas that may not be easily accessible by humans. At the same time, this task often requires recognizing anomalies from a significant distance and/or not previously encountered in the past. In this paper, we propose a novel framework that leverages the advanced capabilities provided by Large Language Models (LLMs) to actively collect information and perform anomaly detection in novel scenes. To this end, we propose an LLM-based model dialogue approach, in which two deep learning models engage in a dialogue to actively control a drone to increase perception and anomaly detection accuracy. 
We conduct our experiments in a high fidelity simulation environment where an LLM is provided with a predetermined set of natural language movement commands mapped into executable code functions. Additionally, we deploy a multimodal Visual Question Answering (VQA) model charged with the task of visual question answering and captioning. By engaging the two models in conversation, the LLM asks exploratory questions while simultaneously flying a drone into different parts of the scene, providing a novel way to implement active perception.  By leveraging LLM's reasoning ability, we output an improved detailed description of the scene going beyond existing static perception approaches. In addition to information gathering, our approach is utilized for anomaly detection  and our results demonstrate the proposed method's effectiveness in informing and alerting about potential hazards.


\end{abstract}
\section{Introduction}
In the last few years, drones have witnessed numerous technological advancements, as well as great commercial exposure for their ability to perform difficult tasks, such as surveillance, anomaly detection, and aerial monitoring in challenging environments. To effectively support these tasks and ensure the efficient and autonomous operation of robots, large informative datasets, e.g., containing drone images, action states, and/or anomalies, were necessary in order to cover every possible scenario that could occur~\cite{MISHRA20201,9219585,9412972}. These approaches primarily focused on collecting a large quantity of data and employing different learning techniques to detect possible anomalies in autonomous drone flying scenarios.

With the major advancements in deep learning across numerous domains, there have been multiple attempts to incorporate these modern, more effective technologies for the sake of enhancing autonomous systems' efficiency and capability. By deploying larger, more advanced deep learning models a substantial improvement in performance was witnessed~\cite{10.2523/IPTC-20111-MS,unlu:hal-02864552}. Nevertheless, these methods lack the ability to \textit{actively perceive} the scene in order to issue the appropriate control commands and further improve the perception accuracy based on the current conditions. Such active perception approaches have shown promising results in other relevant domains in recent years~\cite{bajcsy2018revisiting,saito2021select,manousis2023enabling}. However, it is not trivial to implement such methods in open-world setups. 

\begin{figure}[htbp]
\centerline{\includegraphics[width=0.90\linewidth]{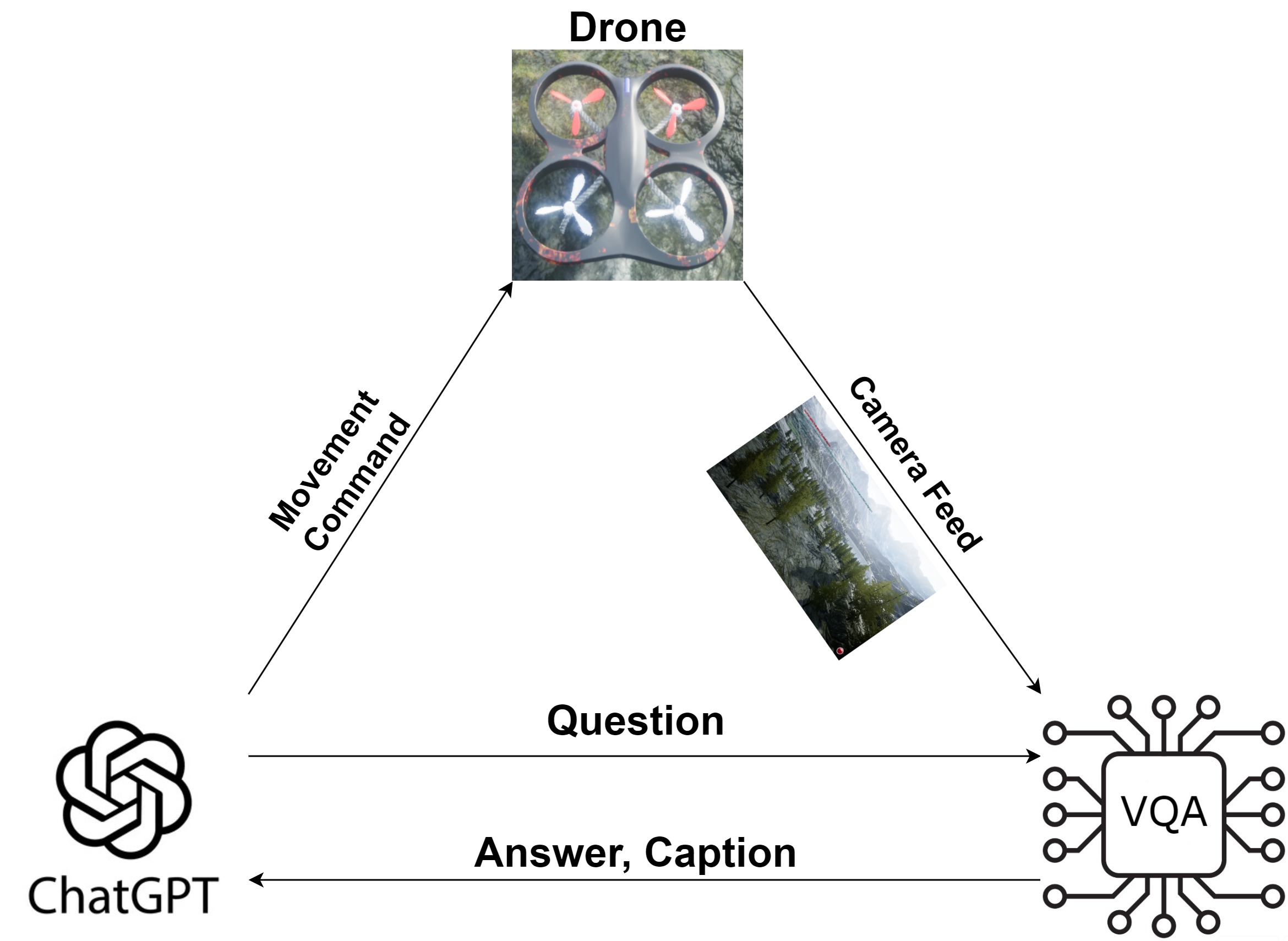}}
\caption{Overview of the proposed model dialogue approach. First a drone captures an image. This image, along with an appropriate question, is fed to the employed VQA model. Then, the VQA model provides a response that is fed to the LLM model which in turn issues a movement command and a new exploratory question.}
\label{fig:outline}
\end{figure}

The main contribution of this paper is a novel approach for active perception and anomaly detection that leverages the capabilities of recent Large Language Models (LLMs) by developing a \textit{model dialogue} approach in which two deep learning models interact in order to continuously improve the final prediction. To this end, we equip the employed LLM with complete navigational control through a set of specific textual commands that ultimately navigate a drone in real time, implementing an active perception scheme in which the drone explores the scene and exploits potential hazardous scenarios and anomalies. Furthermore, we incorporate a Visual Question Answering (VQA) model in order to engage the two models in interactive conversation from which the LLM acts as a controller that can extract meaningful textual information about the unknown scene in which the drone operates. Our goal is to provide a detailed description of the scene gathered throughout the conversation along with explanations that led to these  decisions. This dialogue process leads to an active perception pipeline in which we can  gather additional information about the scene, as well as validate the scene details. The conducted experimental evaluation  shows that the proposed approach can indeed enable a drone to successfully navigate an unknown open environment and provide an explainable and detailed description of the scene in a zero-shot fashion, as well as detect anomalies and output potential safety measures in response to potentially hazardous observations. The code used for the conducted experiments, including detailed prompts and experimental results are provided at \url{https://github.com/Tzoulio/Large_Models_Dialogue_for_Active_Perception}.

The rest of the paper is structured as follows. Section~\ref{sec:related} introduces the related work, while the proposed method is presented in Section~\ref{sec:proposed}. The experimental evaluation is provided in Section~\ref{sec:experimental}, while Section~\ref{sec:conclusions} concludes the paper.

\section{Related Work}
\label{sec:related}

The task of Visual Question Answering \cite{agrawal2016vqa} has  increased in popularity in recent years, with the ability to combine computer vision  with Natural Language Processing (NLP) resulting in a system that can process two types of different modalities at the same time. Such an ability is crucial in robotics applications considering they are often applied to scenarios and environments that require handling such multimodal data. By giving a robot the ability to process multiple data together at once, they increase the quality and quantity of information they acquire, which in turn expands their overall knowledge of the world. As a result, there have been multiple attempts at applying VQA in robotics. Some works focus on having the robot interact with the environment and come up with an answer to a specific question, mimicking the VQA task. Deng et al. \cite{Deng__2021} uses VQA in a robotic manipulation scenario. They train a Deep Q Network (DQN) and through reinforcement learning teach the robot to continuously manipulate objects until they come up with the right answer. In \cite{gordon2018iqa} a Hierarchical Interactive Memory Network (HIMN) was deployed as a controller that allows the system to store and retrieve information hierarchically in the form of memory and enables the robot to provide an answer by interacting with its environment in real-time.  EmbodiedQA \cite{das2017embodied} is another approach that deploys a robot in an unknown environment in which the robot learns to navigate through using imitation learning and ultimately gathers the appropriate information to answer the question.  Our work leverages the recent advances in VQA as a fundamental part of the proposed pipeline by employing a VQA model which acts as the \textit{sensing} model, which processes the data acquired from the world and answers questions regarding these.

After the breakthrough that LLMs made in the field of AI, researchers have been constantly finding ways to utilize them in robotic applications. A lot of works leverage the LLMs' reasoning capabilities and language understanding ability to act as a communicator between the human operator who issues a command in natural language and the robot who executes the command in the form of code \cite{vemprala2023chatgpt,Lamine2023FromWT,ye2023improved,liang2023code}. These approaches either directly map specific commands to code snippets that are applied on the robot directly or provide enough resources to the LLM to construct code and make specific API calls that will produce the correct result on the robot, as specified in the natural language prompt. Generally, a lot of research is focused on advancing the LLM capabilities further, by implementing different modules together with the LLM in an attempt to give it multi-modal capabilities~\cite{wu2023visual,shen2023hugginggpt,wu2023nextgpt}. This resulted in a lot of works which combined multi-modal variations of LLMs into robot task planning \cite{shridhar2021cliport,bucker2022reshaping,stepputtis2020languageconditioned}. These works utilize imitation learning to teach a control agent how to perform the natural language tasks which are learned from a dataset consisting of sets of demonstrations during different timestamps. In other works, such as \cite{vemprala2023chatgpt}, users are able to control an aerial drone through natural language and prompt engineering.  The proposed method goes beyond these approaches by employing a dialogue-based approach, in which only one model has full access to the visual modality and the other model can interact with this model through textual prompts.

The proposed method is more closely related to recent attempts to combine LLMs with VQA models. Some  works \cite{zhu2023chatgpt,rotstein2023fusecap,levy2023chatting,rs16030441} focus on initiating a conversation between the two models to enhance the VQAs ability in the captioning task. They start with a general caption of a query image and through ChatGPT's ability of understanding and generalising textual information an active dialogue between the LLM and the VQA module is initiated. During the dialogue, ChatGPT makes inquiries about possible information that the image might contain. Afterwards, the VQA model answers by confirming or denying and providing additional information for the scene. The process continues until ChatGPT outputs a detailed description containing all the knowledge it gathered through the conversation. Other methods follow a similar approach \cite{hu2023promptcap,yu2023prophet,ravi2022vlcbert} by providing complementary knowledge to the LLM in the form of captions. This enhances the quality and flow of information, resulting in better answers and captions for the query images.  Our method builds on this idea, going beyond these approaches by implementing \textit{active perception} through the drone's navigation scheme. We collect a different image of the scene each time the drone reaches a new position. At the same time, the employed LLM asks an exploratory question with each movement command and the VQA model provides an answer and a caption. This way, we are able to gather more information (extracted by the different captions we get in every position) as well as explore parts of the initial image that the camera could not see either by them being obscured or simply by being too far away.

\section{Proposed Method}
\label{sec:proposed}

\begin{figure*}
    \centering
    \includegraphics[width=1\linewidth, trim={0cm 0cm 0cm 0cm}]{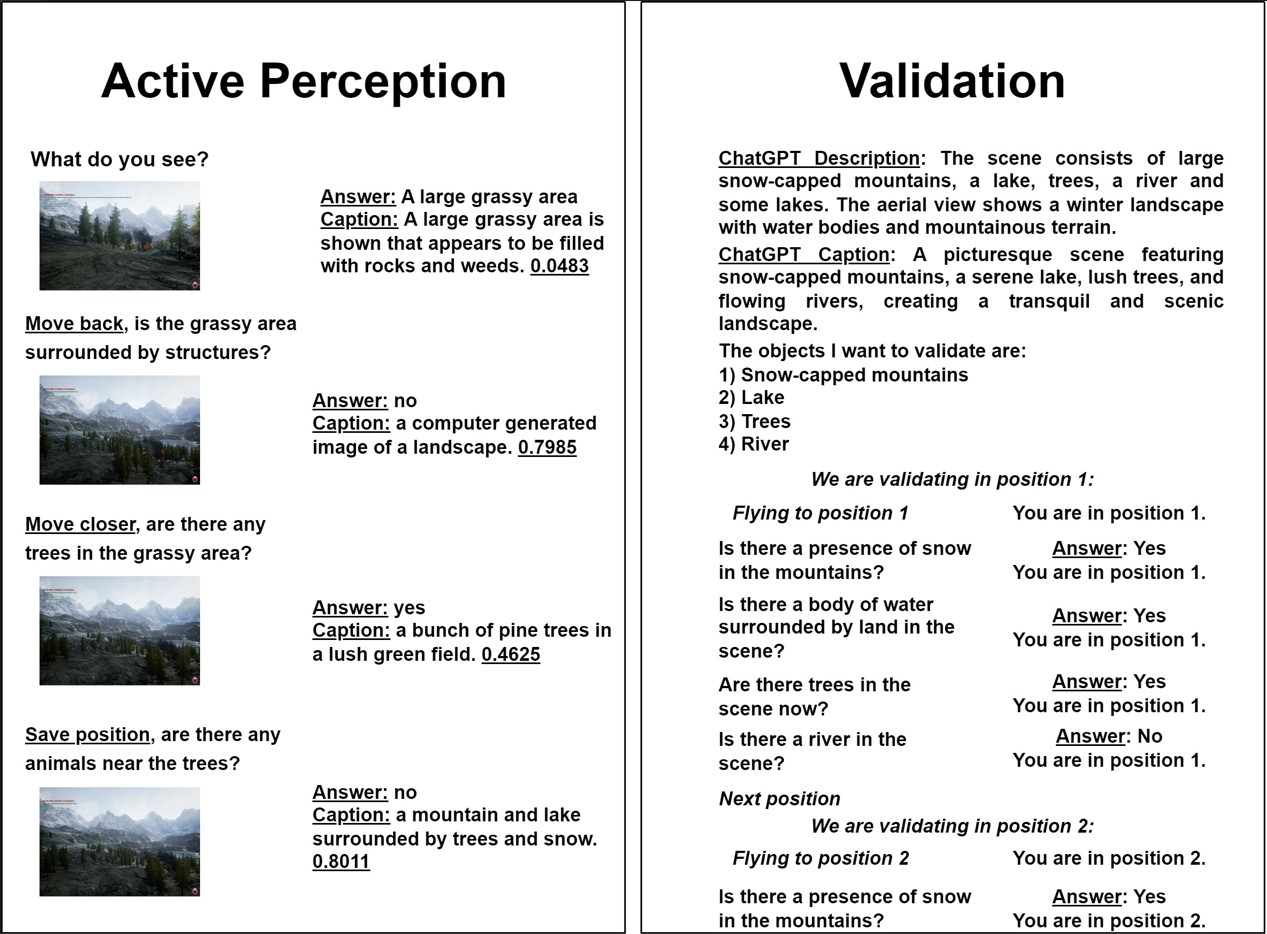}

    \includegraphics[width=0.6\linewidth, trim={84cm 0 0 0}, clip]{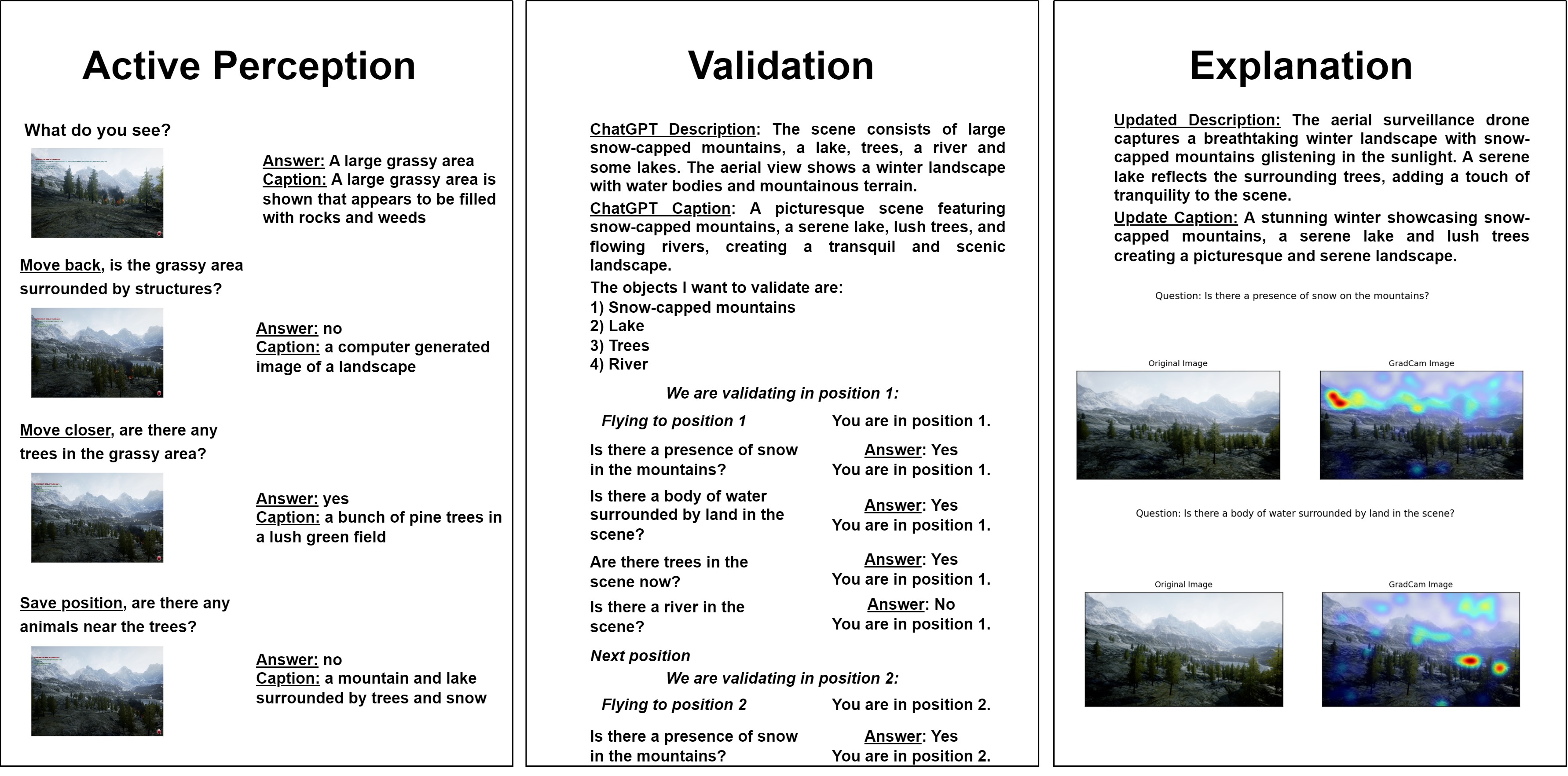}

    \caption{A typical example of the operation of the proposed method. During active perception, the two models engage in a conversation and exchange information. In validation, a premature description and caption are chosen together and information is validated by revisiting the saved positions. Then, in the explanation mode, the final description and caption are provided together with attention maps.}
    \label{fig:example}
\end{figure*}

In this work, we aim to equip a drone with active perception and anomaly detection capabilities in order to provide a robust scene description, as depicted in Fig.~\ref{fig:outline}. First, the drone leverages a VQA model which provides descriptions of the environment through  captions. In this way, the VQA model provides a way for the LLM to ``sense'' the environment through text. Additionally, the VQA model outputs an image-caption matching score in order to help the LLM distinguish between good and bad captions. Then, the LLM validates the gathered textual information through the VQAs question-answering module combined with active perception and ultimately provides a generalized scene description together with explainable attention maps. The outline of the proposed approach is shown through an example in Fig.~\ref{fig:example}. This example should be used as a reference point through the description provided in this Section, since it further clarifies how the proposed method works.

\begin{figure*}
    \centering
    \makebox[\textwidth][c]{\includegraphics[width=0.99\textwidth]{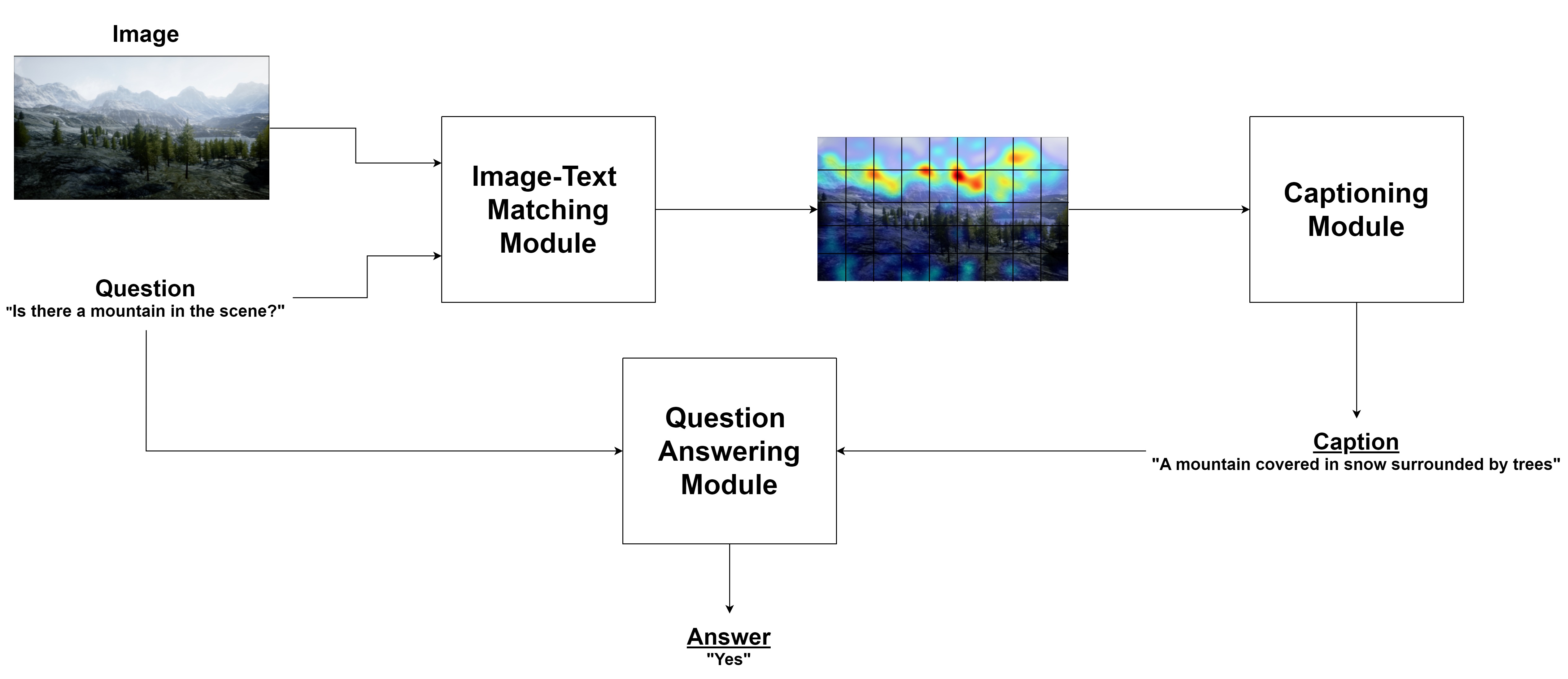}}
    \caption{The employed VQA architecture.}
    \label{fig:vqa}
\end{figure*}

For the VQA model, we incorporate the Plug-and-Play VQA (PnP-VQA) \cite{tiong2023plugandplay} framework, as shown in Fig.~\ref{fig:vqa}. To perform the task of image captioning, image-question pairs are processed by a pre-trained vision-language model called BLIP \cite{li2022blip} which is also able to output a similarity score between the image and the question. The image is split into \textit{K} patches and through GradCAM \cite{8237336}, a feature-attribution interpretability technique, they are able to provide the most relevant image patches. Finally, the image captioning module of BLIP is combined with top-\textit{k} sampling to generate captions only for the relevant patches. Subsequently, the produced caption and question are fed into the question answering module to produce the answer. For the LLM, we employed the GPT3.5 as our model~\cite{brown2020language}.

Let the LLM model denoted by $f(\mathbf{A}, \mathbf{C})$, which takes two distinct text sequences as input $\mathbf{A} = [A_1, A_2, \dots, A_n]$, $\mathbf{C} = [C_1, C_2, \dots, C_m]$  and outputs a response sequence $\mathbf{Q} = [{Q}_1, {Q}_2, \dots, Q_k]$, in the form of a question i.e., $\mathbf{Q} = f(\mathbf{A}, \mathbf{C})$, where $\mathbf{A}$ denotes the answer to a previous question by the VQA model (if exists) and $\mathbf{C}$ denotes a textual description (caption) of the current scene. In this work, we employed the GPT3.5 model to implement $f(\cdot)$, while we feed the concatenated $\mathbf{A}$ and $\mathbf{C}$ to the model.
We assume $A_i$, $C_i$ and $Q_i$  denote the indices of words, while $n$, $m$ and $k$ denote the corresponding sequence lengths. Similarly, the VQA network $g(\mathbf{Q}, \mathbf{I})$ takes as input the output sequence of the LLM $\mathbf{Q}$, as well as an image $\mathbf{I}$, producing two different textual sequences $ \mathbf{A}, \mathbf{C}\ = g(\mathbf{Q},\mathbf{I})$, where $\mathbf{A}$ is the answer to the question and $\mathbf{C}$ denotes the caption for the image. Then, these outputs are fed to the LLM and this process repeats in an iterative fashion. 

To grant the LLM control of the drone we first define a set of diverse functions, each one in charge of a specific navigational output. Afterwards, we provide the drone with a detailed prompt consisting of a set of commands mapped to a specific function apiece, certain rules the GPT3.5 outputs must follow, the general goal of the task and tips on how to filter and extract information from captions.
Additionally, to prevent hallucination~\cite{zhang2023sirens}, i.e., imaginative and fabricated outputs from the controller, we begin the prompt by informing LLM that it is in a game scenario, the commands serve as its controls and the goal is to provide a detailed description of the observed scene while looking out for any possible anomalies that could lead to hazardous situations.
The list of commands is split into: 
\begin{enumerate}[label=(\roman*)]
\item Active perception commands
\begin{enumerate}[label=(\alph*)]
\item Move closer, to move 10 meters forward.
\item Move back, to move 5 meters backwards.
\item Move right, to move 10 meters to the right.
\item Move left, to move 10 meters to the left.
\end{enumerate}

\item General control commands
\begin{enumerate}[label=(\alph*)]
\item Save position, to save the current position of the drone.
\item Ask a question, to ask exploratory questions.
\item I know enough, to return to the starting position.
\end{enumerate}
\end{enumerate}
Additionally, we divide the diverse list of rules the LLM must follow into:
\begin{enumerate}[label=(\roman*)]
    \item General Rules, to make sure  LLM outputs the commands and questions correctly. 
    \item Active Perception rules, which ensure the proper movement of the drone.
    \item Visual Question Answering rules, in order to utilize the captions and answers as efficiently as possible and optimize the procedure.
\end{enumerate}

The propose pipeline consists of the following: an \textit{active perception} mode, a \textit{validation} mode and an \textit{explanation} mode. Throughout active perception mode, the drone's camera takes snapshots of the observed scene and the controller asks questions while simultaneously issuing different movement commands.  The process always starts with the question ``What do you see?''.
Consequently, the VQA model returns an answer, a caption and a percentage indicating if the caption matches the specific image to help distinguish between accurate and inaccurate captions. Through multiple diverse captions from different angles of the scene, the LLM model is able to gain knowledge and by leveraging its language understanding capabilities it is able to generalize and understand the context, as well as output possible safety measures for the specific scene.  Then, during exploration mode, we encourage the LLM (by providing the appropriate prompt) to use the command \textit{save position} whenever it deems it necessary in order to save the current drone position and revisit it during validation mode. The process continues until the LLM uses the command \textit{I know enough} and transitions to validation mode.

During the validation mode, we ask the LLM to output a description and a caption of its current knowledge, along with which parts it wants to validate. We add random Gaussian noise to the saved positions, in order to gain different question-image pairs before inputting them to the VQA model again. In each new position, the controller asks one validating question for each targeted piece of information it wants to validate and we also save the question-image pairs which hold the highest matching score percentage for explanation mode. Afterwards, the controller compiles all the answers in each revisited position and leverages an ensemble approach to update the scene description and caption. In the end, the drone returns to its starting position outputting the final description, caption and the safety rules about the scene.

Finally, in order to provide the ability to {explain} the conclusions drawn by the developed pipeline, we extract the GradCAM's visualization  from our VQA model in order to output attention maps on the validated images, as shown in Fig.~\ref{fig:example}. As a result, when the drone returns to its starting position it is able to output the question-image pairs through an attention mask, highlighting the parts of the image that lead to its decisions on the captioning and question-answering tasks.

\section{Experimental Evaluation}
\label{sec:experimental}

\begin{figure}[h]
\centerline{\includegraphics[scale=0.17]{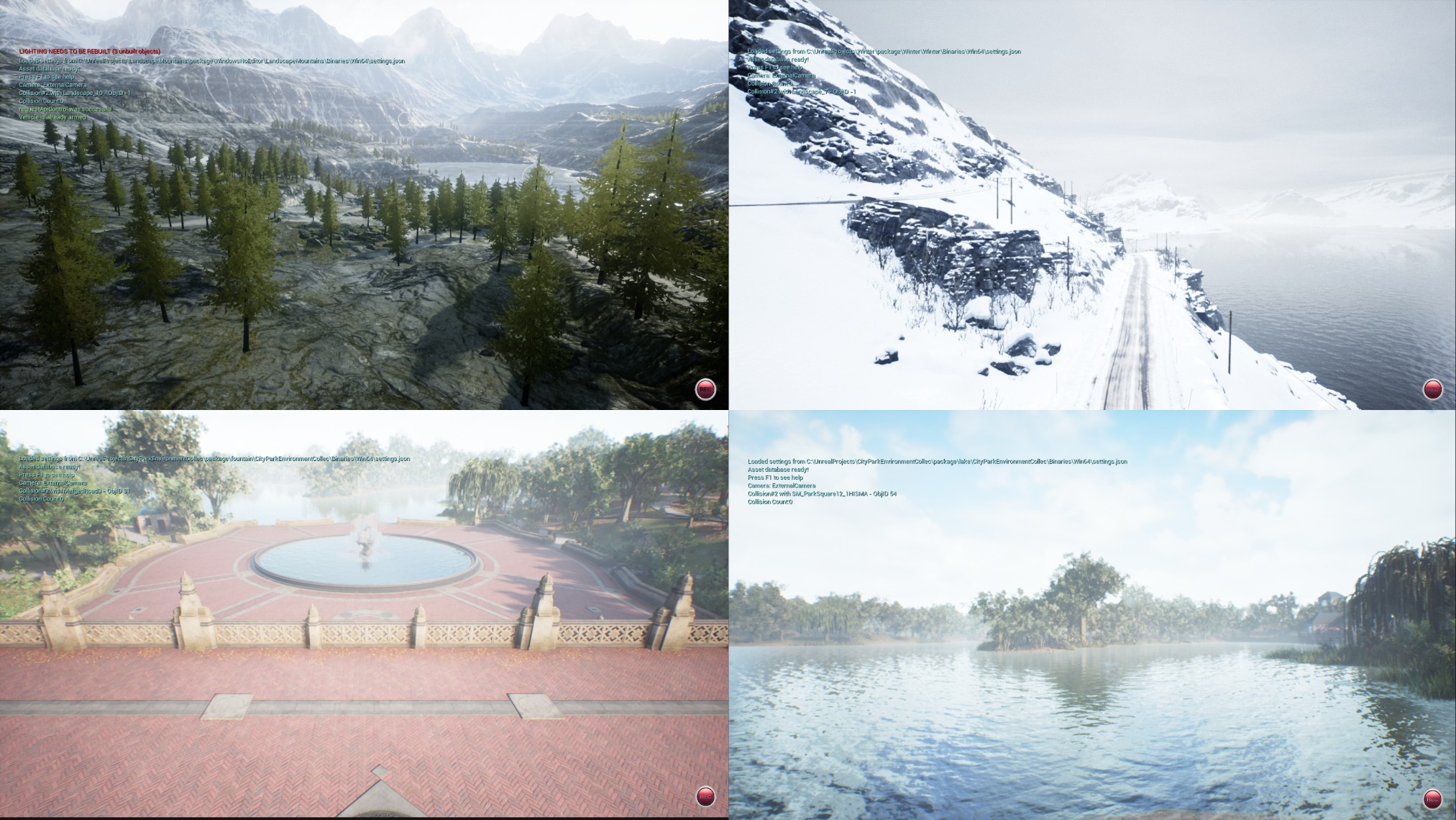}}
\caption{Four different environments were used for the conducted experiments: a  mountain landscape, a snowy road, a public square and a lake.}
\label{fig:typical-environments}
\end{figure}

All the experiments were conducted using the Airsim simulation environment \cite{shah2017airsim}. It is built upon Unreal Engine 4 and consists of a physics engine and different environmental, vehicular and sensory models. By testing out the quad-rotor vehicular model in multiple environments we can simulate a plethora of scenarios that provide physical and visual feedback adjacent to the real world. Specifically, our experiments take place in typical surveillance environments such as a mountain landscape, a lake, a public square and a snowy road, as shown in Fig.~\ref{fig:typical-environments}.

To quantitatively evaluate the performance of the proposed method we compute the caption-image matching score (using the VQA model) at the drone's spawn position and at every subsequent position revisited during the validation module. We then calculate the average caption-image matching score across all positions for ten independent experiments. The results are reported in Table~\ref{table:results}, where we compare the baseline score (directly using the description at the starting position of the drone), and the final validated result of the proposed method (``Proposed'').  Our results indicate that in different environments, the proposed method consistently enhances the caption-image matching score, suggesting that the generated captions provide more relevant information that aligns well with the scene. \hl{Furthermore, we present the average run time required, to obtain a validated, detailed scene description with explainable attention maps. 
Given that the average experiment time is approximately 12 minutes and recognizing that such a duration is impractical in hazardous situations, we introduce a special rule in our prompt. This rule stipulates that whenever the proposed method detects a potential anomaly, it must immediately stop exploration and proceed with validation and result generation. By implementing this rule, we reduce the average experiment time to under 5 minutes in anomaly induced scenarios.}


\begin{table}[h]
\caption{Average image-caption matching score (calculated over ten runs) for each of the employed environments.}

\centering
\begin{tabular}{c|cc|c}
\toprule
\textbf{Environment} & \textbf{Baseline}                 &  \textbf{Proposed} & \textbf{Time of Experiment}\\ 
\midrule
Mountain Landscape            & 0.384 &  \textbf{0.585}     & \textbf{12mins 57secs}     \\ 
Public Square        & 0.361    & \textbf{0.699}      & \textbf{12mins 28secs}    \\ 
Snow road & 0.458  & \textbf{0.629} & \textbf{11mins 48secs}  \\ 
Lake &0.451 & \textbf{0.690}  & \textbf{13mins 26secs} \\ \bottomrule
\end{tabular}

\label{table:results}
\end{table}

\newpage

\begin{figure}[h]
\centerline{\includegraphics[scale=0.16]{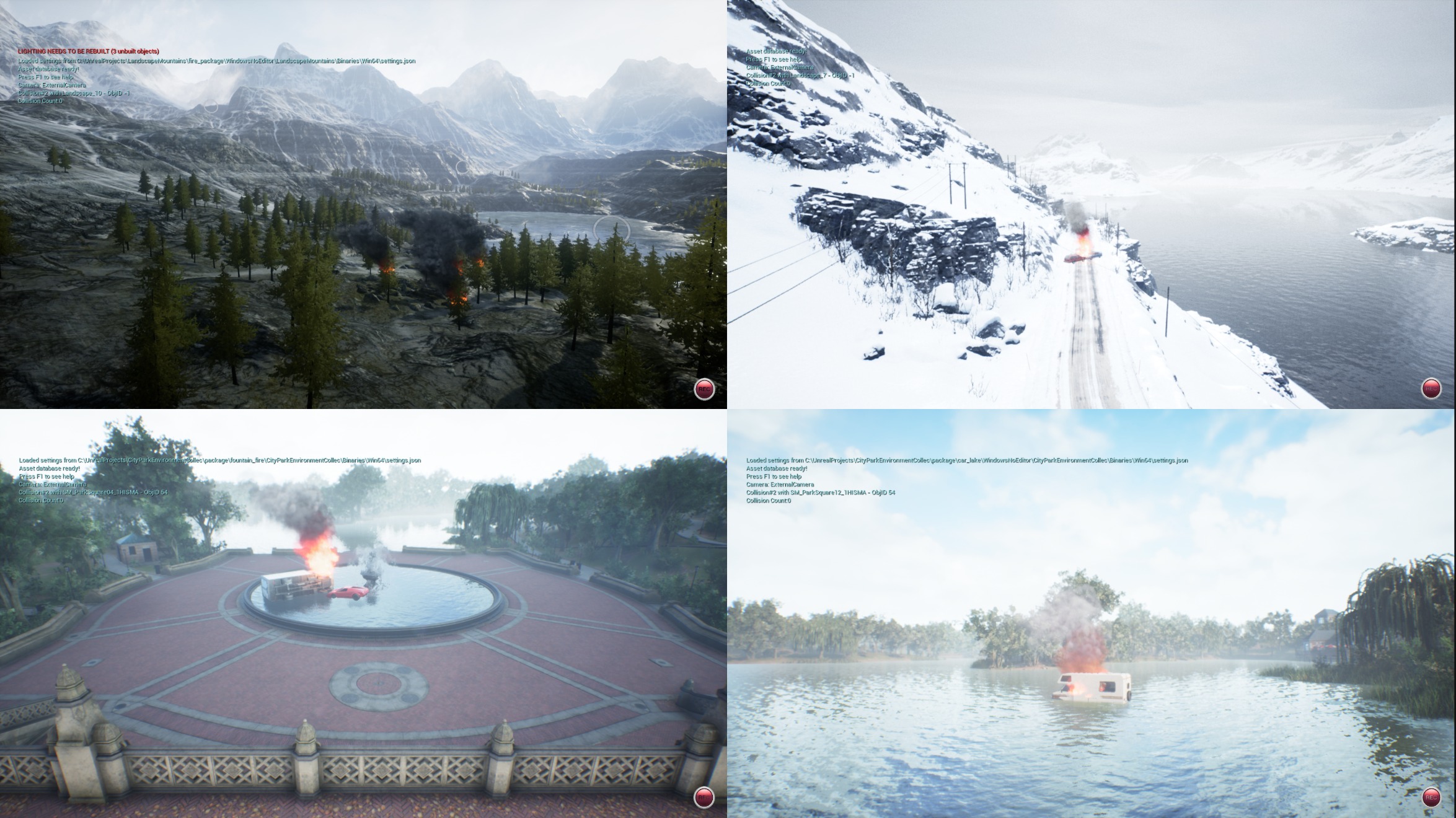}}
\caption{Example anomalies in the four different environments. Note that some anomalies are challenging to detect and require very careful inspection of the input frame.}
\label{fig:example_anomalies}
\end{figure}

Additionally, we assess our system's performance on the task of anomaly detection. By introducing potential hazards or dangerous elements, such as fires and car crashes, into each scene (refer to Fig.~\ref{fig:example_anomalies} for some example anomalies), we evaluate the baseline framework's ability to accurately identify anomalies, comparing it with the performance of our proposed system following the active perception and validation phases. We consider the system successful in anomaly detection when it identifies the anomaly in its captions in a coherent and grammatically logical manner. To  evaluate the proposed method  in scenes that contain anomalies, we deploy hazards in three distinct scenarios. Initially, we position a potential hazard within the range of the drone's spawn point. Subsequently, we increase the distance between the drone's spawn point and the hazard. Finally, we place the hazard in an obscured view from the initial drone spawn point necessitating movement to locate it. We conduct the experiments ten times for each environment and present the accuracy of anomaly detection (averaging the ten runs over the three setups), comparing the baseline and the proposed method, in Table~\ref{table:anomaly}.

\begin{table}[!ht]
\caption{Comparing anomaly detection accuracy between baseline and the proposed method.}
\label{table:anomaly}
\centering
\begin{tabular}{c|c|c}
\toprule
\textbf{Method} & \textbf{Environment} & \textbf{Anomaly Detection Score}\\
\midrule
Baseline & {Mountain Landscape} & 0.53\\
Proposed & {Mountain Landscape} & \textbf{0.90}\\
\midrule
Baseline & Public Square &  0.43\\
Proposed & Public Square &  \textbf{0.73}\\
\midrule
Baseline & Lake & 0.26 \\
Proposed & Lake & \textbf{0.76} \\
\midrule
Baseline & Snow & 0.20 \\
Proposed & Snow & \textbf{0.83} \\
\bottomrule
\end{tabular}
\end{table}

\begin{figure}[!h]
\centerline{\includegraphics[scale=0.40]{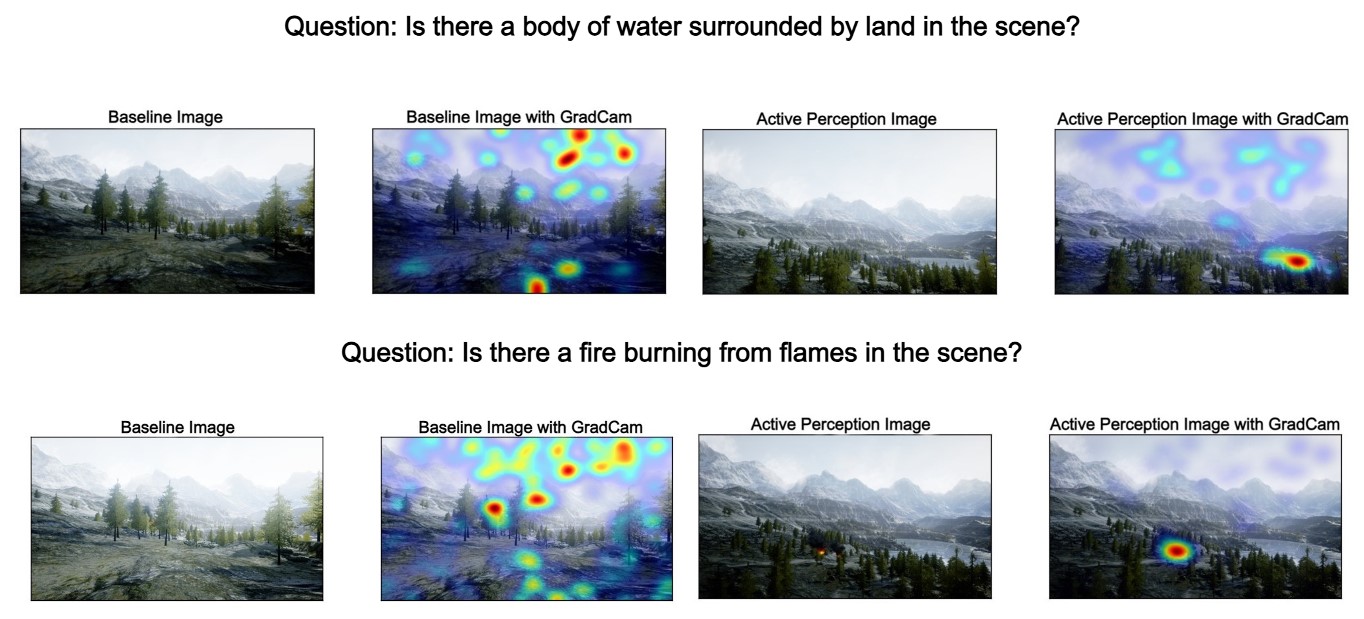}}
\caption{Two examples for two different questions, indicating the additional explain ability capabilities that can be provided by the proposed pipeline.}
\label{fig:explainability}
\end{figure}

These results indicate that the drone succeeded in providing a description and caption about the unknown scene whilst only relying on outputs from the VQA model in the form of text. Moreover, when hazardous anomalies are introduced, altering the scene to an unsafe condition, our system successfully identifies the danger and suggests necessary safety precautions. Finally, the proposed pipeline can also provide interpretable attention maps, leveraging GradCAM's capabilities, both for the intermediate and final questions/captions, which showcase the validated information in order for a human operator to assess. Two indicative examples are shown in Fig.~\ref{fig:explainability}, highlighting the improved explainability capabilities provided by the proposed method. \hl{Furthermore, in Table~\ref{table:example_caption}, we compare the captions provided by the baseline model with the captions provided by the proposed framework and in Table~\ref{table:description} we provide the detailed scene descriptions leveraged by our proposed framework. Note that in most cases the proposed method leads to a more accurate description. However, hallucinations can still occur despite the validation process. Increasing the number of examination points and/or adding additional validation steps could help further reduce these occurrences. }

\begin{table}[!ht]
\caption{Caption examples provided by the baseline VQA model and the proposed model. We highlight the correct pieces of information with the color \textcolor{ForestGreen}{green}, the wrong ones with the color \textcolor{red}{red} and the ambiguous ones with \textcolor{YellowOrange}{orange}.}
\label{table:example_caption}
\begin{tabularx}{\textwidth}{@{} P{30mm} |@{} P{45mm} |@{} P{45mm} }
    \toprule
\textbf{Scene}
&   \textbf{Baseline}   &   \textbf{Proposed }                                                \\
    \midrule
\textbf{Mountain Landscape}
      & A view of \textcolor{ForestGreen}{rocky mountain peaks} that looks into the horizon
      &   A serene \textcolor{ForestGreen}{mountainous landscape with mist, snow-capped mountains, and trees}.     \\
\midrule

\textbf{Snowy road in mountainside}
       & The \textcolor{ForestGreen}{snowy mountain} is \textcolor{ForestGreen}{covered in a thick blanket of snow}.
       &   A \textcolor{ForestGreen}{snowy mountain with a road leading into glacier water}. \\
\midrule

\textbf{Public Square}
  & \textcolor{ForestGreen}{A fountain park} \textcolor{YellowOrange}{filled with lots of water}.
        &   A lively \textcolor{ForestGreen}{fountain park shrouded in dense fog with water shoots} creating a mysterious atmosphere \\
\midrule
\textbf{Lake}  & A group of \textcolor{ForestGreen}{tall vegetation} \textcolor{YellowOrange}{on a river}.
        &   A tranquil \textcolor{ForestGreen}{lake setting} with \textcolor{red}{ducks}, \textcolor{ForestGreen}{tall vegetation, and lush green plants}, offering a picturesque natural landscape. \\
\midrule
\midrule

\textbf{Mountain Landscape with fire} 
        & A huge \textcolor{ForestGreen}{flame} and a cloud of \textcolor{ForestGreen}{black smoke}.
        &   A devastating \textcolor{ForestGreen}{forest fire consumes the valley, threatening the green vegetation and trees in its path}.                       \\
\midrule



\midrule
\textbf{Lake with a car fire}
  & The \textcolor{red}{steam rises} in the clouds on a foggy day.
       &   \textcolor{ForestGreen}{A car crash has occured, with a truck damaged after crashing into a river emitting smoke}, \textcolor{red}{individuals trying to move the stuck truck}. \\
   
    \bottomrule
\end{tabularx}

\end{table}


\begin{table}[!ht]
\caption{We showcase our methods ability to provide descriptions of the scenes, after the information was gathered through Active Perception and after it was validated through our validation module. We highlight the correct pieces of information with the color \textcolor{ForestGreen}{green}, the wrong ones with the color \textcolor{red}{red} and the ambiguous ones with \textcolor{YellowOrange}{orange}.}
\label{table:description}
\scriptsize
\makegapedcells
\setlength\tabcolsep{4pt}
\begin{tabularx}{\linewidth}{@{} P{21mm} |L @{}}
    \toprule
\thead[lb]{Environment}
        &   \thead{Proposed Final Description}                                                \\
    \midrule
\hl{\textbf{Mountain Landscape}}
        &   The aerial surveillance drone has captured \textcolor{ForestGreen}{a serene mountain landscape with trees covering its slopes}. While there is \textcolor{ForestGreen}{no visible forest in the scene}, \textcolor{ForestGreen}{a clear lake} adds to the natural beauty of the surroundings. The \textcolor{ForestGreen}{absence of human activity} enhances the peacefulness of the environment.     \\
        \hline
\textbf{Mountain Landscape with fire}
        &   The aerial surveillance drone captures a dramatic scene with \textcolor{ForestGreen}{a group of mountains featuring rocky peaks in the background}. In the foreground, \textcolor{ForestGreen}{a fire rages} with \textcolor{red}{red lava} and \textcolor{ForestGreen}{flames}, casting a fiery glow. On the left side, \textcolor{ForestGreen}{a majestic mountain stands tall}, adding to the rugged landscape. Meanwhile, on the right side, another \textcolor{ForestGreen}{fire burns with smoke billowing into the sky}. \textcolor{red}{The background displays a computer artwork}, adding a surreal touch to the overall view.                         \\
        \hline
\textbf{Snowy road in mountainside}
        &   The scene depicts a tranquil \textcolor{ForestGreen}{snowy landscape with no specific objects or anomalies present}. The serene setting is characterized by the peacefulness of the \textcolor{ForestGreen}{snow-covered terrain} and the absence of any notable features. \\
        \hline
\textbf{Snowy road in mountainside with car crash}
        &   The scene depicts \textcolor{ForestGreen}{a snowy road with a truck} traveling on it. The road is \textcolor{ForestGreen}{covered in snow}, and there is \textcolor{ForestGreen}{a mountain nearby covered in heavy snow}. The presence of the truck on the snowy road indicates \textcolor{ForestGreen}{a potential hazardous situation} that needs to be approached with caution. \\
        \hline
\textbf{Public Square}
        &   The scene features \textcolor{ForestGreen}{a round, red tiled courtyard enveloped} in fog, creating an eerie and mysterious atmosphere. The fog obscures the surroundings, adding to the sense of obscurity and intrigue. \textcolor{ForestGreen}{The digital object, previously mentioned, is no longer present in the scene} leaving behind a solitary and enigmatic courtyard. \\
        \hline
\textbf{Public Square with fire}
        &   The scene features \textcolor{YellowOrange}{a small fountain with water spraying, and an outdoor fountain with a fire display}, and \textcolor{red}{a fire torch made of metal}. Both \textcolor{ForestGreen}{the small fountain and fire display} have been confirmed to be present in the scene. The \textcolor{red}{fire torch made of metal is also part of the scene}, adding to the overall ambiance. \\
        \hline
\textbf{Lake}
        &   The scene portrays \textcolor{YellowOrange}{a tranquil river} flowing with ripples at its center. Along \textcolor{YellowOrange}{the riverbank}, \textcolor{ForestGreen}{the trees stand tall and healthy}, framing the water's edge \textcolor{ForestGreen}{without any nearby structures} interrupting the natural beauty. Across \textcolor{YellowOrange}{the river} lies \textcolor{ForestGreen}{a park merging into a dense forest}, enhancing the scene's idyllic charm. A blanket of fog envelops the surroundings, lending an air of mystery and serenity to the landscape. \\
        \hline
\textbf{Lake with fire}
        &   The scene features \textcolor{ForestGreen}{a body of water} with \textcolor{red}{a small boat floating in the middle}. In front of \textcolor{red}{the boat}, \textcolor{ForestGreen}{a tree is engulfed in flames, emitting orange burning flames}. \textcolor{ForestGreen}{The fire has spread to the bush tucker on a field with trees}. However, \textcolor{ForestGreen}{there is no floating island engulfed by flames} as previously mentioned. \textcolor{ForestGreen}{Smoke rises from the burning objects, creating a hazardous environment.} \\
   
    \bottomrule
\end{tabularx}

    \end{table}
\newpage

\section{Conclusion}
\label{sec:conclusions}

In this paper, we presented a novel framework that employs LLMs to actively collect information and detect anomalies, even in unprecedented situations. We propose a method where two deep learning models engage in dialogue to control a drone and improve anomaly detection accuracy. We test our approach in a realistic simulation environment, where the LLM follows natural language commands to move the drone, while a VQA model answers questions about images. By combining these models, the LLM asks questions while guiding the drone through the scene, providing a unique way to improve perception accuracy, as well as detect potential anomalies. At the same time, by leveraging the explainability capabilities of the employed VQA model, the proposed method can also further improve the explainability of the perception process. \hl{By providing four different types of scenes, with different hazardous situations in them and without requiring any fine-tuning or retraining of the models, we demonstrate the potential of the proposed method for handling open-ended adaptation in-the-wild. Additionally, to the best of our knowledge, there is currently no other established way to implement and evaluate active perception in unstructured open-world setups. Therefore, this work opens several research directions, including effective evaluation of approaches that extend beyond static perception and pave the way for applications in other areas as well.}

\subsubsection{Acknowledgements} 
The work presented here has been partially supported by the RoboSAPIENS project funded by the European Commission’s Horizon Europe programme under grant agreement number 101133807. This publication reflects the authors’ views only. The European Commission is not responsible for any use that may be made of the information it contains.

\bibliographystyle{IEEEtran}
\bibliography{IEEEabrv, main.bib}

\end{document}